\documentclass[]{spie}  %>>> use for US letter paper
\usepackage{amsmath,amsfonts,amssymb}
\usepackage{graphicx}
\usepackage{float}

\usepackage{array}
\usepackage{longtable}

\usepackage[colorlinks=true, allcolors=blue]{hyperref}
\usepackage{multirow}

\title{Enhanced Landmark Detection Model in Pelvic Fluoroscopy using 2D/3D Registration Loss}
\author[1,2,4,*]{Chou Mo}
\author[1,2,3,*]{Yehyun Suh}
\author[5]{J. Ryan Martin}
\author[1,2,3,\dag]{Daniel Moyer}
\affil[1]{Department of Computer Science, Vanderbilt University, Nashville, TN, USA}
\affil[2]{Vanderbilt Institute for Surgery and Engineering, Nashville, TN, USA}
\affil[3]{Vanderbilt Lab for Immersive AI Translation, Nashville, TN, USA}
\affil[4]{Department of Mathematics, University of California-Los Angeles, Los Angeles, CA, USA}
\affil[5]{Department of Orthopedic Surgery, Vanderbilt University Medical Center, Nashville, TN, USA}

\authorinfo{*Equal Contribution, \dag Corresponding Author, Daniel Moyer: daniel.moyer@vanderbilt.edu}

\begin{document} 
\maketitle

\begin{abstract}
Automated landmark detection offers an efficient approach for medical professionals to understand patient anatomic structure and positioning using intra-operative imaging.
While current detection methods for pelvic fluoroscopy demonstrate promising accuracy, most assume a fixed Antero-Posterior view of the pelvis.
However, orientation often deviates from this standard view, either due to repositioning of the imaging unit or of the target structure itself.
To address this limitation, we propose a novel framework that incorporates 2D/3D landmark registration into the training of a U-Net landmark prediction model. We analyze the performance difference by comparing landmark detection accuracy between the baseline U-Net, U-Net trained with Pose Estimation Loss, and U-Net fine-tuned with Pose Estimation Loss under realistic intra-operative conditions where patient pose is variable.
\end{abstract}

% Include a list of keywords after the abstract 

\section{Introduction}
\label{sec:intro}  % \label{} allows reference to this section

Anatomical landmark detection is the process of identifying clinically significant points on medical images that correspond to specific anatomical structures in the underlying physical anatomy. The landmarks in radiographic images function as reference points for surgical planning, post-operative assessment, and other biomechanical analyses \cite{moon, Mulford, Suh23}. This is particularly significant in orthopedic procedures for the pelvic region, where complex three-dimensional (3D) bone structure is often interpreted from two-dimensional (2D) projections. Landmarking on the pelvic bone presents difficulties due to its wide anatomical differences between patients, overlapping bone structures in projected radiographs, and the irregular geometry of key structures such as the acetabulum and sacroiliac joints.
\par
The accurate localization and pose estimation of the pelvis is important for Total Hip Arthroplasty (THA), as it guides surgeons in proper component placement and achieving the optimal biomechanical alignment. The acetabular rim serves as a reference for cup placement, while the femoral head center helps determine hip joint mechanics, and the greater trochanter assists with soft tissue balancing. Further, the assessment of gait and loading patterns, together with developmental dysplasia evaluation and trauma surgery, requires pelvic landmarks as crucial references. The traditional method of radiographic landmark annotation through manual marking requires extensive expertise and yields significant observer variability \cite{chan2025development}, which could impact both surgical outcomes and research reliability.
\par
2D/3D registration is the process of finding dense correspondences between a preoperative 3D volume, commonly collected from CT scans, with an intraoperative 2D image, such as fluoroscopy or X-rays. This is usually parameterized by an explicit rigid transformation; the objective of 2D/3D registration amounts to estimating the camera pose $\theta \in SE(3)$, which is a 6-degrees of freedom transformation (three rotational variables and three translational variables) \cite{gao2020generalizing, unberath2021impact, Gopalakrishnan_2024_CVPR}. In the surgical context, the ``camera pose'' represents the 3D orientation of the imaging device relative to the patient’s anatomy (though in fluroscopy this is actually the X-ray source and detector). Aligning and projecting an entire 3D volume is computationally consuming, which is why anatomical landmarks are adopted as correspondence points for 2D/3D registration. With 3D landmarks and camera pose that illustrate the 3D orientation in which the 2D radiograph is taken, models can more accurately locate the 2D position of landmarks on any designated radiograph of that same 3D volume. 
\par
Method proposed in Suh et al. 2023 \cite{suh2023dilationerosion} uses a U-Net \cite{UNet} structure to detect landmarks, which treat the landmarking problem as a semantic segmentation task. In this work, a probability map is generated from the trained U-Net, and landmarks are extracted based on the maximum probability of a landmark being located at a particular pixel location. While this approach has been proven effective for segmentation tasks in other fields, it is limited in precisely locating landmarks. Existing training methods adopt pixel-wise cross-entropy loss between the output probability maps and ground truth masks. This method does not directly penalize spatial inaccuracies in landmark locating - i.e., a network can achieve a low segmentation loss but produce landmarks that are pixels away from their true anatomical locations. Meanwhile, more intuitively, it may be more efficient to prioritize the geometric distance error between the predicted and ground truth landmark coordinates in a landmark detection model. 
\par
To address these limitations, we propose integrating pose estimation loss into the landmark detection framework. Using pose estimation loss, the model explicitly accounts for the geometric error between the predicted 2D landmark coordinates and their ground truth coordinates. Our approach recognizes that accurate landmark detection and camera pose estimation are coupled problems: small deviations in 2D predictions can lead to significant pose errors. Therefore, instead of basing optimization solely on probability map loss, we introduce a pose estimation loss component that directly penalizes the Euclidean distance error. We experimentally show that fine-tuning with pose estimation loss improves landmark detection accuracy. In this study, we compared three model structures: a U-Net trained exclusively with pose estimation loss, a U-Net trained with a composite loss function (a combination of segmentation loss with weighted pose estimation loss), and a U-Net that is initially trained with segmentation loss and subsequently fine-tuned using pose estimation loss.

% \begin{table}[h]
% \centering
% \begin{tabular}{|l|c|c|c|c|c|}
% \hline
% \multicolumn{6}{|c|}{\textbf{Validation Set (SDR(\%))}} \\
% \hline
% \textbf{Experiment} & Mean RMSE & \textbf{$<2$} & \textbf{$<3$} & \textbf{$<5$} & \textbf{$<10$} \\
% \hline
% Baseline U-Net & 5.58 & 8.49 & 17.96 & 41.35 & 80.83 \\
% \hline
% U-Net Trained with Only Pose Estimation Loss  & 11.73 & 1.97 & 4.61 & 12.34 & 39.64 \\
% \hline
% U-Net Trained with Pose Estimation Loss  & 11.73 & 1.97 & 4.61 & 12.34 & 39.64 \\
% \hline
% U-Net Fine-Tuned with Pose Estimation Loss & \textbf{5.085} & 6.36 & 14.05 & 33.69 & 73.06 \\
% \hline
% \end{tabular}
% \label{tab:landmark_performance}
% \end{table}

\section{Method}
\subsection{Landmark Prediction Model}
% \subsubsection{Landmark Extraction from U-Net Output}
Given an input image $I$, the U-Net outputs a heatmap $h_i(x,y)$ for each of the landmarks $i \in \{1, ..., 8\}$, representing the likelihood of the $i_{th}$ landmark located at the pixel $(x,y)$. We adopt Soft-Argmax as a differentiable method to extracting the 2D landmarks from heatmaps:
\begin{equation}
\label{eq:softargmax}
\hat{\mathbf{l}}_i = (\hat{x}_i, \hat{y}_i) = 
\left(
\sum_{x=0}^{W-1} \sum_{y=0}^{H-1} x \cdot \text{softmax}(h_i)_{(x,y)}, \ 
\sum_{x=0}^{W-1} \sum_{y=0}^{H-1} y \cdot \text{softmax}(h_i)_{(x,y)}
\right)
\end{equation}
\begin{equation}
\text{softmax}(h_i)_{(x,y)} = 
\frac{\exp(h_i(x, y)/\tau)}{\sum_{x', y'} \exp(h_i(x', y')/\tau)}
\end{equation}
Where $W$ and $H$ denote the width and height of the heatmap, respectively, $\tau$ is the temperature parameter that controls the sharpness of the softmax distribution, and $(x', y')$ are dummy indices used to normalize the softmax over all spatial locations in the heatmap.

\subsection{Camera Pose and Landmark-based 2D/3D Registration}
\label{sec:Camera}

We define the orientation of the medical imaging camera as a rigid body transformation with 6 degrees of freedom. The camera pose $\theta$ consists of three rotation parameters $R \in SO(3) $ and three translation parameters $t \in \mathbb{R}^3$. The rotations $R$ are parameterized by $(r_x, r_y, r_z)$, representing an Euler angle rotation around the three axes, while the translations are parameterized by $(t_x, t_y, t_z)$, representing the displacement from the origin, which we define to be the center of the CT volume. 
\par
By convention, landmark-based 2D/3D registration is computed by extracting the ground truth 3D coordinates of landmarks $L_{GT} \in {\mathcal{M}_{8 \times 3}(\mathbb{R})}$ on the CT and the predicted 2D coordinates of landmarks $l_{pred} \in {\mathcal{M}_{8 \times 2}(\mathbb{R})}$ that are output from U-Net, then finding the pose $\theta$ that aligns the projection of the 3D coordinates and the 2D coordinates \cite{grupp2020automatic}. We conduct the registration using a limited memory Broyden-Fletcher-Goldfarb-Shanno (L-BFGS) PyTorch optimizer 
\cite{liu}, aiming to minimize the loss function:
\begin{equation}
\theta^* = \arg\min_{\theta} \mathcal{L}(\theta).
\end{equation}
We define the loss function $\mathcal{L}(\theta)$ as a sum of squared residuals: 
\begin{equation}
\label{eq:2d3dreg}
\mathcal{L}(\theta) = \sum_{i=1}^{N} \left\| \hat{\mathbf{l}}_i(\theta) - \mathbf{l}_i \right\|^2
\end{equation}
where $\theta = (r_x, r_y, r_z, t_x, t_y, t_z) \in \mathbb{R}^6$ is the 6-DoF pose parameter vector, $\hat{\mathbf{l}}_i(\theta)$ is the predicted 2D landmarks of the $i$-th 3D landmark under pose $\theta$, $\mathbf{l}_i$ is the U-Net outputted 2D landmarks, and $N$ is the number of landmarks. The residual is computed via our perspective projection calculations, which align with DiffDRR projections (see the appendix for more details).

\subsection{Pose Estimation}
We solve for the optimal pose $\theta^*$ with gradient-based optimization, given predicted 2D landmark coordinates from U-Net $\mathbf{l}_i \in \mathcal{M}_{8\times2}(\mathbb{R})$ and the ground truth 3D landmark coordinates from the CT volume $L_{GT} \in \mathcal{M}_{8\times3}(\mathbb{R})$. As described in \hyperref[sec:Camera]{Section 2.2}, we minimize the loss function defined in Equation \eqref{eq:2d3dreg}.

The rotations $R \in SO(3)$ are parameterized using the Rodrigues formula. The formula maps a 3-dimensional rotation vector $\mathbf{r} = (r_x, r_y, r_z) \in \mathbb{R}^3$ to a rotation matrix. The magnitude $\|\mathbf{r}\|$ denotes the rotation angle, while the direction denotes the rotation axis. The rotation matrix is defined as:
\begin{equation}
R = I + \sin(\theta) \cdot K + (1 - \cos(\theta)) \cdot K^2
\end{equation}
where $\theta = \|\mathbf{r}\|$, $I \in \mathbb{R}^{3\times3}$ is the identity matrix, and $K$ is the skew-symmetric matrix constructed from the unit axis $\hat{\mathbf{r}} = \mathbf{r}/\theta$:
\begin{equation}
K = \begin{bmatrix} 0 & -\hat{r}_z & \hat{r}_y \\ \hat{r}_z & 0 & -\hat{r}_x \\ -\hat{r}_y & \hat{r}_x & 0 \end{bmatrix}
\end{equation}
This parameterization ensures that $R$ fulfills the orthogonality constraint $R^T R = I$ with only three parameters.

In the perspective projection calculation \cite{suh20252d}, $\hat{\mathbf{l}}_i(\theta)$ transforms each 3D landmark $\mathbf{p}_i \in L_{GT}$ to its corresponding 2D image coordinates. The camera intrinsic matrix $K \in \mathbb{R}^{3\times3}$ includes information about the source-to-detector distance (SDD) and principal point. The projection is computed as the following:
\begin{equation}
\mathbf{p}'_i = R \mathbf{p}_i + \mathbf{t}
\end{equation}
\begin{equation}
\hat{\mathbf{l}}_i = \frac{1}{z'_i} \begin{bmatrix} K_{11} x'_i + K_{13} z'_i \\ K_{22} y'_i + K_{23} z'_i \end{bmatrix}
\end{equation}
where $\mathbf{p}'_i = (x'_i, y'_i, z'_i)^T$ are the transformed 3D coordinates. Further, the $z'_i$ are clamped to a minimum value of $10^{-3}$ before performing the perspective division for numerical stability.

We initialize the optimization with $\mathbf{r}^{(0)} = \mathbf{0}$ and $\mathbf{t}^{(0)} = \mathbf{0}$, corresponding to the identity transformation. At each iteration, the loss $\mathcal{L}(\theta)$ from Equation \eqref{eq:2d3dreg} is computed via forward projection, while the gradients $\partial \mathcal{L}/\partial \mathbf{r}$ and $\partial \mathcal{L}/\partial \mathbf{t}$ are computed through automatic differentiation. We adopt the Adam optimizer with learning rate $\eta = 10^{-3}$ to update the pose parameters. To prevent numerical instability, we constrain $\mathbf{r} \in [-2\pi, 2\pi]$ and $\mathbf{t}$ within physically reasonable bounds (as specified in \hyperref[sec:Data]{Section 3.1}). The optimization ends after a maximum of 100 iterations, yielding the estimated pose $\hat\theta = (\hat{\mathbf{r}}, \hat{\mathbf{t}})$.

\subsection{Pose Estimation Loss based Landmark Prediction Model}
   \begin{figure}[t]
   \begin{center}
   \includegraphics[width=\linewidth]{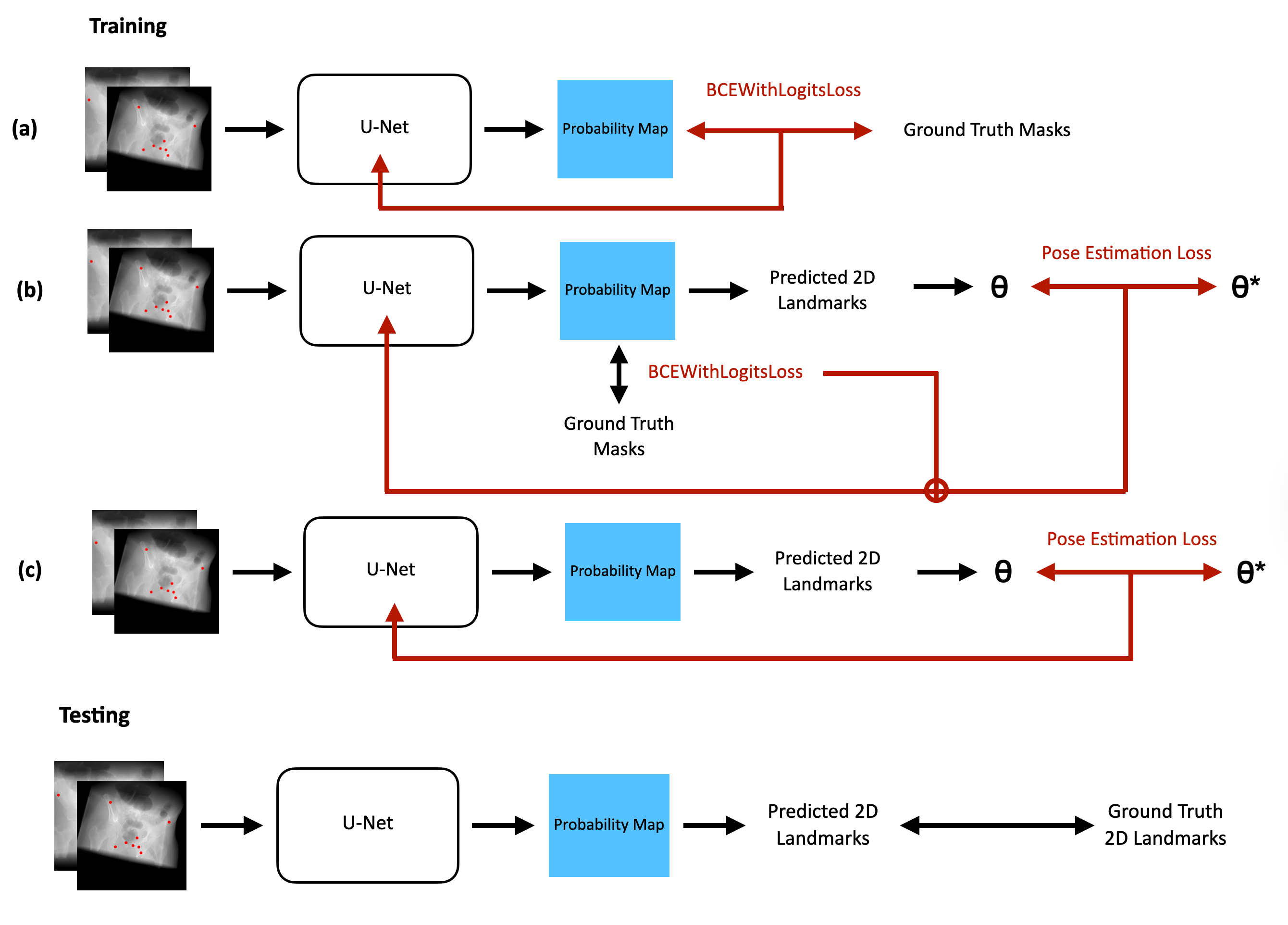}
   \end{center}
   \caption[ModelDiagram] 
   { \label{fig:ModelDiagram} 
Model pipeline diagram for three tested models. From top to bottom: (a) Baseline U-Net, (b) U-Net trained on pose estimation loss, and (c) U-Net fine-tuned on pose estimation loss. $\theta$ denotes the predicted pose, and $\theta^*$ denotes the ground truth pose.}
   \end{figure} 

We evaluated three training strategies for landmark detection using a U-Net architecture: \textbf{Baseline U-Net}, \textbf{ U-Net trained on pose estimation loss}, and \textbf{U-Net fine-tuned on pose estimation loss}. Each of the three pipelines adopts a different loss mechanism to optimize landmark detection accuracy. The model pipeline architectures and testing process are illustrated in Figure~\ref{fig:ModelDiagram}.
\par
The \textbf{Baseline U-Net} (Figure~\ref{fig:ModelDiagram}a) employs a standard U-Net architecture trained with segmentation loss. During training, the model is optimized using Binary Cross-Entropy with Logits Loss (BCEWithLogitsLoss), evaluating the loss between the U-Net probability heatmap and the ground truth mask. The second approach, \textbf{ U-Net trained on pose estimation loss}, (Figure~\ref{fig:ModelDiagram}b) introduces a composite pose estimation loss component during training. The composite pose estimation loss function is formulated as:
\begin{equation}
\mathcal{L}_{composite} = \mathcal{L}_{seg}+ \lambda \cdot \mathcal{L}_{pose}
\end{equation}
where $\lambda$ is the pose loss weighting factor. The pose estimation loss is computed as:
\begin{equation}
   \mathcal{L}_{pose} = \sqrt{\frac{1}{6} \sum_{j=1}^{6} (\theta_j - \theta_j^{*})^2}
\end{equation}
The pose estimation loss component requires converting the predicted heatmaps to 2D landmark coordinates using differentiable soft-argmax (as shown in Equation~\ref{eq:softargmax}), followed by 2D/3D registration to estimate the 6 degrees of freedom pose (as shown in Equation~\ref{eq:2d3dreg}).
\par
The \textbf{U-Net fine-tuned on pose estimation loss} (Figure~\ref{fig:ModelDiagram}c) adopts a two-stage training strategy: In the first stage, the U-Net is trained using only segmentation loss as in the baseline model. In the second stage, fine-tuning is conducted on the pre-trained model using only the pose estimation loss. Similar to the \textbf{ U-Net trained on pose estimation loss}, pose estimation in fine-tuning is calculated via 2D/3D registration with the L-BFGS optimizer (see Equation~\ref{eq:2d3dreg}). 

\section{Experiments}
\subsection{Data and Experimental Setting}
\label{sec:Data}
   \begin{figure}  [ht]
   \begin{center}
   \includegraphics[height=5cm]{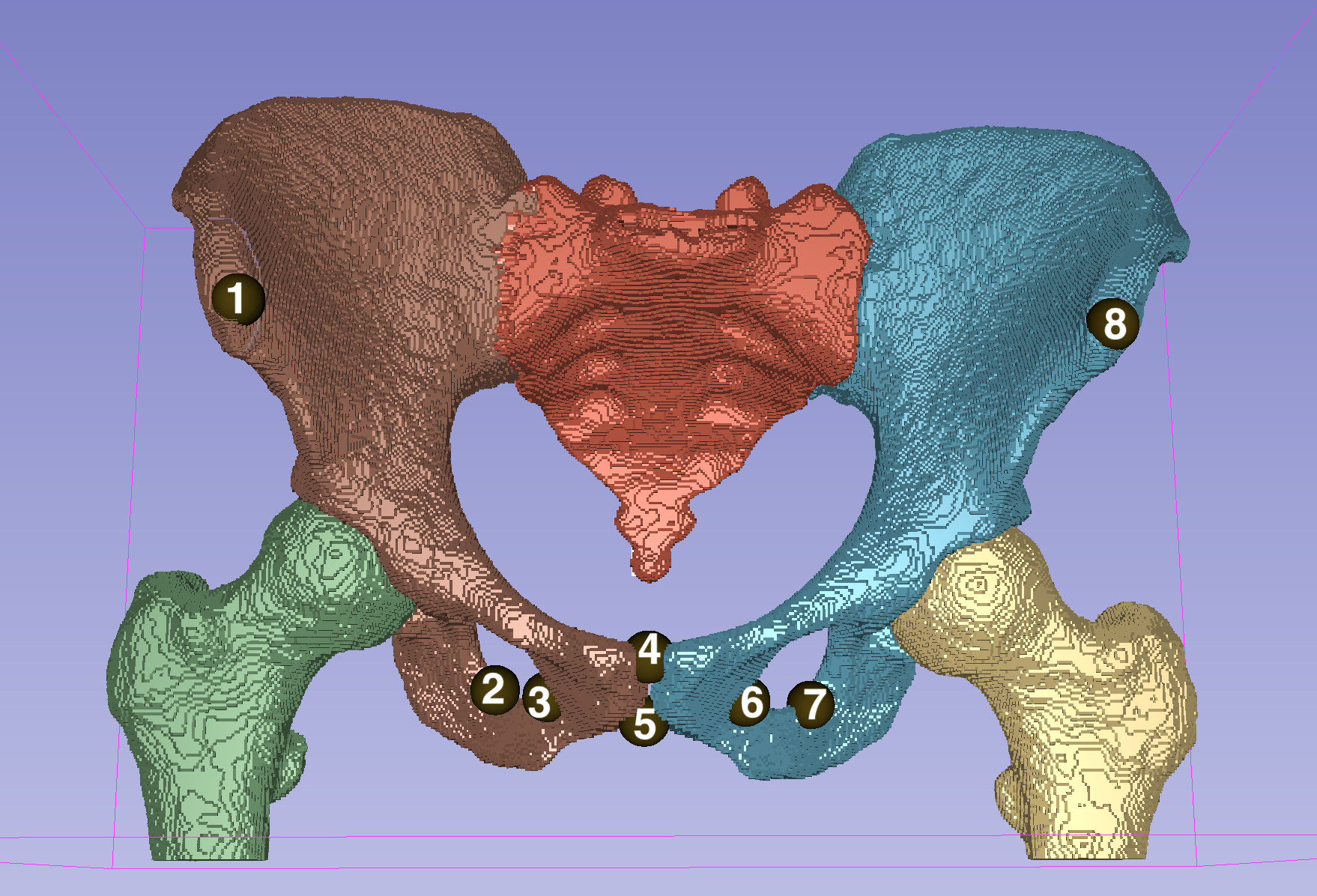}
   \end{center}
   \caption[landmarkplacement] 
   { \label{fig:landmarkplacement} 
Landmark placement of eight anatomical landmarks on the pelvic bone, enlarged for visualization.}
   \end{figure} 
% A total of 8 anatomical landmarks on the pelvic bone were adopted, following the placement defined by Grupp, et al \cite{Grupp20}. The landmarks are annotated via Slicer on 3D CT scans. As shown in Fig.~\ref{fig:landmarkplacement}, Landmarks 1 and 8 are the left and right anterior superior iliac spine (ASIS), 2 and 7 are the left and right inferior obturator foramen (IOF), 3 and 6 are the left and right medial obturator foramen (MOF), 4 is the superior pubis symphysis (SPS), while 5 is the inferior pubis symphysis (IPS). 

% We gathered ninety CT scans for our training and testing dataset from the CTPEL Database \cite{ctpel19}. Eight landmarks were manually placed on each CT scan via Slicer. From the dataset, eighty scans were used for training all three models, and ten were used for evaluation. One-hundred simulated X-ray images were generated for each CT scan using DiffDRR \cite{Gopalakrishnan22}. The source-to-detector distance was set to 1020 mm, and the image size $768 \times 768$. All images are then rescaled to $512 \times 512$ prior to training. 
We gathered ninety CT scans for our training and testing dataset from the CTPEL Database \cite{ctpel19}. Eight landmarks were manually placed on each CT scan via Slicer \cite{pieper20043d}, following the placement defined by Grupp, et al \cite{grupp2020automatic}. As shown in Fig.~\ref{fig:landmarkplacement}, Landmarks 1 and 8 are the left and right anterior superior iliac spine (ASIS), 2 and 7 are the left and right inferior obturator foramen (IOF), 3 and 6 are the left and right medial obturator foramen (MOF), 4 is the superior pubis symphysis (SPS), while 5 is the inferior pubis symphysis (IPS). From the dataset, eighty scans were used for training all three models, and ten were used for evaluation. One-hundred simulated X-ray images were generated for each CT scan using DiffDRR \cite{Gopalakrishnan22}. The source-to-detector distance was set to 1020 mm, and the image size $768 \times 768$. All images are then rescaled to $512 \times 512$ prior to training. 
\par
Random rotations $(r_x, r_y, r_z)$ within the range of $[-45, 45]$ degrees and random translations $(t_x, t_y, t_z)$ within the range of $[-50, 50]$ were applied for image and landmark projection using DiffDRR. Before passing into pose estimation, loaded 3D voxel coordinates are transformed and centered to real-world coordinates through $L_{GT} = (L_{voxel} - C) \times S_{CT}$. $C$ is the center of the CT volume, and $S_{CT}$ denotes the pixel spacing of the CT scan in x, y, and z direction. Then, an additional y-direction displacement of the volume-to-detector distance is added to the $L_{GT}$ coordinates. Further, the extracted predicted 2D coordinates $\hat{\mathbf{l}}$ also undergo several transformations prior to pose estimation: 
\\
\indent(1) Scale adjustment for X-ray image rescaling: $\hat{\mathbf{l}} \cdot \frac{768}{512}$ \\
\indent(2) X-value centering: $\hat{\mathbf{l}}[:, 0] - \frac{768}{2}$ \\
\indent(3) Y-value inversion and centering:
   $-(\hat{\mathbf{l}}[:, 1] - \frac{768}{2})$ \\
%For more details on how we aligned perspective projection with DiffDRR projections, see the appendix. 
\par
All three models are trained with a learning rate $1\times10^{-4}$, and the seed is set to $42$. The \textbf{Baseline U-Net} is trained for 28 epochs, with encoder depth $5$, the \textbf{U-Net trained on pose estimation loss} is trained for 28 epochs, and the \textbf{U-Net fine-tuned on pose estimation loss} is fine-tuned for one epoch.
\subsection{Evaluation}
All three models are evaluated through the same pipeline. As shown in the bottom section of Figure~\ref{fig:ModelDiagram}, predicted 2D landmark coordinates are extracted from probability heatmaps using Argmax, which are then compared with the ground truth 2D landmarks to arrive at a Root Mean Square Error (RMSE) across all landmarks and per landmark. The ground truth 2D landmarks were extracted from the DiffDRR X-ray images during dataset generation. Each model was evaluated on 1600 images from unseen views from the eighty training CT scans (``Novel View'' dataset) and on 1000 images for the ten evaluation CT scans (``Novel Subject'' dataset). 

\section{Results}
\begin{table}
\centering
\resizebox{\textwidth}{!}{%
\begin{tabular}{|p{1.5cm}|l|c|c|c|c|c|c|c|c|c|}
\hline
\multirow{2}{*}{Dataset} & \multirow{2}{*}{Model} & \multicolumn{9}{c|}{RMSE (pixels)} \\
\cline{3-11}
 & & Mean & L1 & L2 & L3 & L4 & L5 & L6 & L7 & L8 \\
\hline
\multirow{4}{1.5cm}{\textbf{Novel View}} & Baseline U-Net & 8.58 & 11.49 & \textbf{4.71} & \textbf{12.34} & \textbf{16.66} & 5.43 & \textbf{6.87} & 6.00 & 5.17 \\
\cline{2-11}
 & U-Net Trained w/PEL & 20.82 & 39.95 & 13.83 & 28.34 & 19.51 & 15.36 & 14.02 & 20.00 & 15.54 \\
\cline{2-11}
 & U-Net Trained w/Only PEL & DIV & - & - & - & - & - & - & - & - \\
\cline{2-11}
 & U-Net Fine-Tuned w/PEL & \textbf{8.45} & \textbf{10.27} & 6.67 & 14.02 & 17.75 & \textbf{3.67} & 7.18 & \textbf{4.12} & \textbf{3.89} \\
\hline
\multirow{4}{1.5cm}{\textbf{Novel Subject}} & Baseline U-Net & 5.58 & \textbf{5.68} & 8.33 & \textbf{4.68} & 4.01 & 8.65 & \textbf{4.07} & \textbf{3.07} & 6.12 \\
\cline{2-11}
 & U-Net Trained w/PEL & 11.73 & 11.63 & 23.33 & 10.59 & 6.15 & 10.38 & 7.7 & 10.46 & 13.57 \\
\cline{2-11}
 & U-Net Trained w/Only PEL & DIV & - & - & - & - & - & - & - & - \\
\cline{2-11}
 & U-Net Fine-Tuned w/PEL & \textbf{5.09} & 7.45 & \textbf{8.01} & \textbf{4.68} & \textbf{2.03} & \textbf{2.03} & 7.52 & 4.48 & \textbf{4.48} \\
\hline
\end{tabular}%
}
\caption[RMSEcomparison] 
{ \label{tab:RMSEcomparison} 
RMSE comparison of different approaches' performance on the novel view set and the held-out subject set. Pose Estimation Loss is abbreviated as PEL.}
\end{table}

   \begin{figure}
   \begin{center}
   \includegraphics[width=0.9\linewidth]{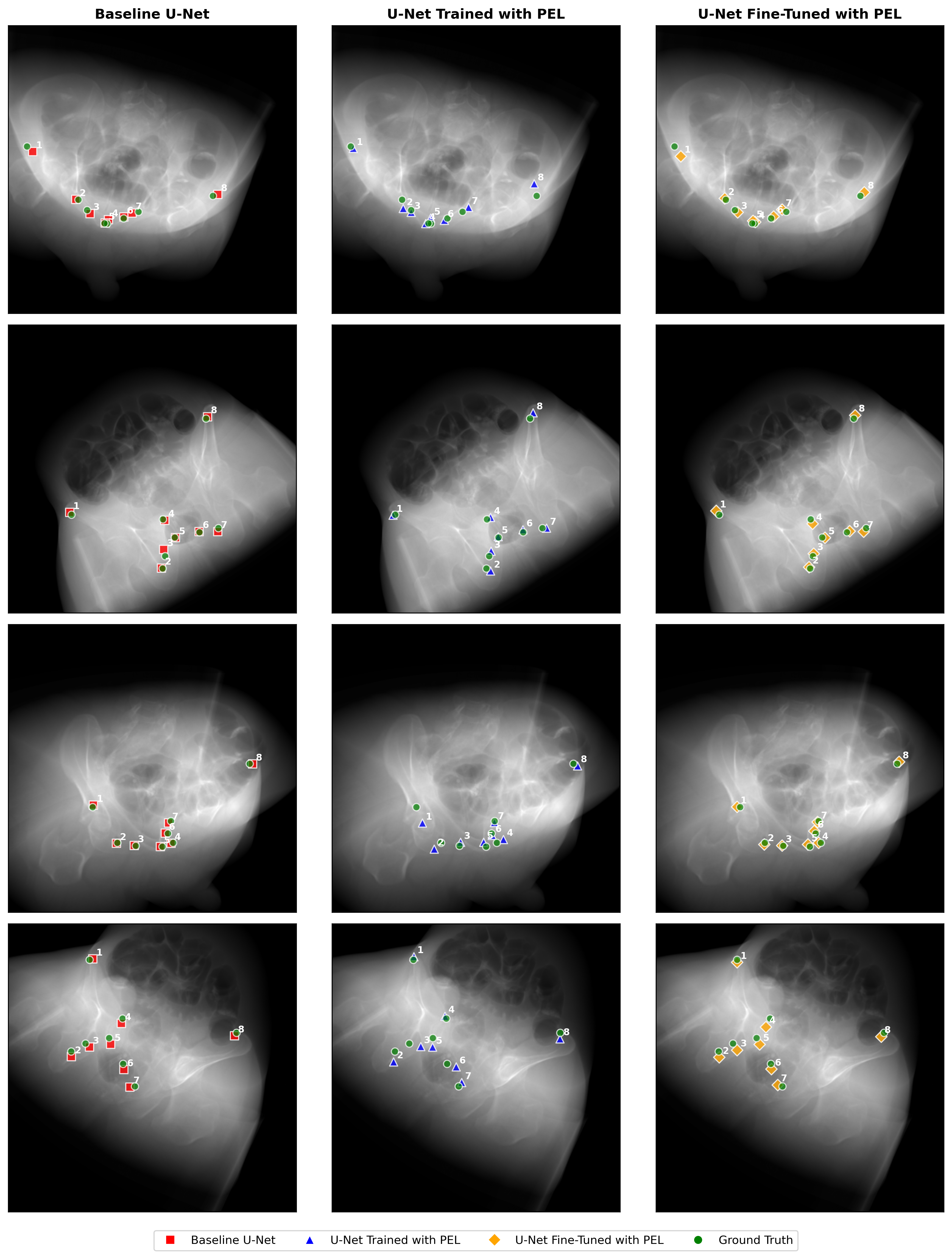}
   \end{center}
   \caption[ModelDiagram] 
   { \label{fig:ModelDiagram} 
Model pipeline diagram for three tested models. From top to bottom: (a) Baseline U-Net, (b) U-Net trained on pose estimation loss, and (c) U-Net fine-tuned on pose estimation loss. $\theta$ denotes the predicted pose, and $\theta^*$ denotes the ground truth pose.}
   \end{figure} 
As shown in Table~\ref{tab:RMSEcomparison}, overall, the proposed fine-tuned U-Net with (composite) pose estimation loss (PEL) achieved improved landmark detection accuracy compared to the baseline U-Net and trained U-Net on PEL. All models performed better in the external test dataset - images that the models had not seen before - than in the internal test dataset. 
\par
The baseline U-Net achieved mean RMSE values of 8.58 mm and 5.58 mm on internal and external datasets, respectively. The U-Net fine-tuned with PEL showed consistent improvements, achieving 8.45 mm ($1.5\%$ improvement) and 5.09 mm ($8.8\%$ improvement) on the respective datasets. 
\par
The U-Net trained with composite PEL and with only PEL both showed a significant performance decline. Specifically, the U-Net trained with composite PEL had approximately 2.4 and 2.1 times increases in error relative to the baseline. This deterioration may be explained by the difference in optimization goals ingrained in the composite loss function. The segmentation loss component encourages the generation of smooth, blob-like probability distributions appropriate for semantic segmentation. Whereas the PEL concurrently guides the network towards sharp, point-like distributions that reduce geometric registration error. The U-Net trained exclusively using PEL failed to converge during training, leading to divergence (shown as "DIV" in Table~\ref{tab:RMSEcomparison}). This finding suggests that pose estimation loss alone does not provide adequate gradient structure for efficient landmark detection learning from random initialization, as the 2D/3D registration component requires sufficiently precise landmark predictions to yield significant supervision signals.
\par
In comparison to the internal test set, the external test set results demonstrate greater accuracy across all methods. The baseline U-Net achieved a 35\% reduction in RMSE on external data (5.58mm compared to 8.58mm), while the fine-tuned method demonstrated a 40\% improvement (5.09mm versus 8.45mm). This unexpected generalization capacity indicates that the models effectively adjusted to various imaging conditions or patient demographics present in the external dataset.

\section{Discussion}
Although our proposed fine-tuning method yields promising results in enhancing landmark detection accuracy, several limitations require consideration.  First, we adopted DiffDRR to generate simulated X-ray images for our evaluation. These images may not fully reflect the noise, artifacts, and variations in imaging that are present in real intraoperative fluoroscopy \cite{Paalvast}. The effectiveness of our method on clinical images has yet to be confirmed \cite{Vivek}. The present study focused on eight anatomical landmarks on the pelvic bone. Expanding this methodology to a broader array of landmarks or alternative anatomical regions may uncover different optimization results. A further limitation is the extra processing power needed for the L-BFGS optimization to operate in the training loop.  Solving a nonlinear optimization problem is needed for each forward pass, which makes training take longer than with standard segmentation methods.  

Future research may address these constraints via various approaches.  First, validating on actual clinical fluoroscopy images with manually annotated ground truth would furnish more robust evidence for clinical relevance.  Second, exploring different optimization algorithms or pose estimation methods based on learning could reduce the computational cost while maintaining consistent geometry.  Third, exploring adaptive loss weighting schemes that automatically balance the goals of segmentation and pose estimation during training may help us overcome the optimization problems observed in our composite loss experiments.  Lastly, adding temporal consistency to this framework for video fluoroscopy sequences could make it even easier to track landmarks during dynamic surgical procedures.

\section{Conclusion}
This study revealed that the integration of 2D/3D registration constraints through pose estimation loss can improve landmark detection accuracy when implemented through sequential fine-tuning instead of joint optimization. By fine-tuning the baseline U-Net for one epoch, the fine-tuned model achieved higher registration accuracy in internal and external test sets. 
The declined performance of composite loss training and pose estimation loss only methods underscores the significance of suitable training curricula for medical imaging tasks. The sequential method effectively maintains spatial understanding from segmentation training while incorporating geometric constraints for improved localization precision.
These findings establish a viable framework for incorporating 3D anatomical constraints into 2D landmark detection, illustrating that geometric consistency may be efficiently utilized to improve clinical accuracy through better optimization additions.

\acknowledgments % equivalent to \section*{ACKNOWLEDGMENTS}       
This work was supported in part by NSF 2321684 and a VISE Seed Grant.

% References
\bibliography{report} % bibliography data in report.bib

@article{Suh23,
	author = "Yehyun Suh and Peter Chan and J. Ryan Martin and Daniel Moyer",
	title  = "Label Augmentation Method for Medical Landmark Detection in Hip Radiograph Images",
	journal= "arXiv preprint arXiv:2309.16066",
	year   = "2023",
	month  = "September",
	note   = "\url{https://arxiv.org/abs/2309.16066}"
}

@article{moon,
  author = {Moon, Ki-Ryum and Shi Sub Byon and Sung Hyun Kim and Lee, Byoung-Dai},
  month = {04},
  pages = {e29677-e29677},
  publisher = {Elsevier BV},
  title = {Automated assessment of pelvic radiographs using deep learning: A reliable diagnostic tool for pelvic malalignment},
  doi = {10.1016/j.heliyon.2024.e29677},
  volume = {10},
  year = {2024},
  journal = {Heliyon}
}

@article{liu,
  author = {Liu, Dong C. and Nocedal, Jorge},
  month = {08},
  pages = {503-528},
  title = {On the limited memory BFGS method for large scale optimization},
  doi = {10.1007/bf01589116},
  volume = {45},
  year = {1989},
  journal = {Mathematical Programming}
}

@article{Mulford,
  author = {Mulford, Kellen L and Johnson, Quinn J and Mujahed, Tala and Khosravi, Bardia and Rouzrokh, Pouria and Mickley, John P and Taunton, Michael J and Wyles, Cody C},
  month = {10},
  pages = {2024-2031.e1},
  title = {A Deep Learning Tool for Automated Landmark Annotation on Hip and Pelvis Radiographs},
  doi = {10.1016/j.arth.2023.05.036},
  url = {https://pubmed.ncbi.nlm.nih.gov/37236288/},
  volume = {38},
  year = {2023},
  journal = {The Journal of arthroplasty}
}

@article{Paalvast,
  author = {Paalvast, Olivier T. and Hertgers, Omar and Sevenster, Merlijn and Lamb, Hildo J.},
  month = {01},
  title = {Assessing the Image Quality of Digitally Reconstructed Radiographs from Chest CT},
  doi = {10.1007/s10278-025-01406-9},
  urldate = {2025-04-01},
  year = {2025},
  journal = {Journal of Imaging Informatics in Medicine}
}

@misc{Vivek,
  author = {Gopalakrishnan, Vivek and Dey, Neel and Golland, Polina},
  month = {12},
  title = {Intraoperative 2D/3D Image Registration via Differentiable X-ray Rendering},
  url = {https://arxiv.org/html/2312.06358v1},
  urldate = {2025-11-24},
  year = {2023},
  organization = {Arxiv.org}
}

@misc{UNet,
      title={U-Net: Convolutional Networks for Biomedical Image Segmentation}, 
      author={Olaf Ronneberger and Philipp Fischer and Thomas Brox},
      year={2015},
      eprint={1505.04597},
      archivePrefix={arXiv},
      primaryClass={cs.CV},
      url={https://arxiv.org/abs/1505.04597}, 
}

@inproceedings{Gopalakrishnan22,
  title={Fast auto-differentiable digitally reconstructed radiographs for solving inverse problems in intraoperative imaging},
  author={Gopalakrishnan, Vivek and Golland, Polina},
  booktitle={Workshop on Clinical Image-Based Procedures},
  pages={1--11},
  year={2022},
  organization={Springer},
  doi={10.1007/978-3-031-23179-7_1},
  url={https://arxiv.org/abs/2208.12737}
}

@misc{ctpel19,
  author       = { Bryan Connolly and Chunliang Wang },
  title        = { Segmented CT pelvis scans with annotated anatomical landmarks },
  year         = 2019,
  publisher    = { AIDA },
  doi          = { 10.23698/aida/ctpel },
  howpublished = { Available: AIDA Data Hub },
  url          = { https://datahub.aida.scilifelab.se/10.23698/aida/ctpel }
}

@article{chan2025development,
  title={Development of a Deep Learning Model for Automating Implant Position in Total Hip Arthroplasty},
  author={Chan, Peter YW and Baker, Courtney E and Suh, Yehyun and Moyer, Daniel and Martin, J Ryan},
  journal={The Journal of Arthroplasty},
  year={2025},
  publisher={Elsevier}
}

@inproceedings{
suh2023dilationerosion,
title={Dilation-Erosion Methods for Radiograph Annotation in Total Knee Replacement},
author={Yehyun Suh and Aleksander Mika and J. Ryan Martin and Daniel Moyer},
booktitle={Medical Imaging with Deep Learning, short paper track},
year={2023},
url={https://openreview.net/forum?id=bVC9bi_-t7Y}
}

@article{unberath2021impact,
  title={The impact of machine learning on 2d/3d registration for image-guided interventions: A systematic review and perspective},
  author={Unberath, Mathias and Gao, Cong and Hu, Yicheng and Judish, Max and Taylor, Russell H and Armand, Mehran and Grupp, Robert},
  journal={Frontiers in Robotics and AI},
  volume={8},
  pages={716007},
  year={2021},
  publisher={Frontiers Media SA}
}

@InProceedings{Gopalakrishnan_2024_CVPR,
    author    = {Gopalakrishnan, Vivek and Dey, Neel and Golland, Polina},
    title     = {Intraoperative 2D/3D Image Registration via Differentiable X-ray Rendering},
    booktitle = {Proceedings of the IEEE/CVF Conference on Computer Vision and Pattern Recognition (CVPR)},
    month     = {June},
    year      = {2024},
    pages     = {11662-11672}
}

@inproceedings{gao2020generalizing,
  title={Generalizing spatial transformers to projective geometry with applications to 2D/3D registration},
  author={Gao, Cong and Liu, Xingtong and Gu, Wenhao and Killeen, Benjamin and Armand, Mehran and Taylor, Russell and Unberath, Mathias},
  booktitle={International Conference on Medical Image Computing and Computer-Assisted Intervention},
  pages={329--339},
  year={2020},
  organization={Springer}
}

@article{grupp2020automatic,
  title={Automatic annotation of hip anatomy in fluoroscopy for robust and efficient 2D/3D registration},
  author={Grupp, Robert B and Unberath, Mathias and Gao, Cong and Hegeman, Rachel A and Murphy, Ryan J and Alexander, Clayton P and Otake, Yoshito and McArthur, Benjamin A and Armand, Mehran and Taylor, Russell H},
  journal={International journal of computer assisted radiology and surgery},
  volume={15},
  number={5},
  pages={759--769},
  year={2020},
  publisher={Springer}
}

@inproceedings{suh20252d,
  title={2D/3D Registration of Acetabular Hip Implants Under Perspective Projection and Fully Differentiable Ellipse Fitting},
  author={Suh, Yehyun and Martin, J Ryan and Moyer, Daniel},
  booktitle={Workshop on Clinical Image-Based Procedures},
  pages={1--10},
  year={2025},
  organization={Springer}
}

@inproceedings{pieper20043d,
  title={3D Slicer},
  author={Pieper, Steve and Halle, Michael and Kikinis, Ron},
  booktitle={2004 2nd IEEE international symposium on biomedical imaging: nano to macro (IEEE Cat No. 04EX821)},
  pages={632--635},
  year={2004},
  organization={IEEE}
}
\bibliographystyle{spiebib} % makes bibtex use spiebib.bst

\end{document}